\documentclass[twoside,11pt]{article}

\usepackage{jmlrutils}
\usepackage[preprint]{jmlr2e}

\usepackage{lastpage}
\jmlrheading{}{2026}{}{}{}{}{J. Kongmanee and S. Thanapattheerakul}

\ShortHeadings{The Latent Diagnostic Taxonomy}{J. Kongmanee and S. Thanapattheerakul}
\firstpageno{1}

\begin{document}

\title{The Latent Diagnostic Taxonomy: A Framework for Constructing Classifiers and Diagnosing Their Decisions, Applied to Prompt Injection Detection}

\author{\name {Jaturong Kongmanee}\thanks{Work done under Smile Thanapattheerakul’s leadership.} \email {jaturong\_kongmanee@trendmicro.com}
       \AND
       \name {Smile Thanapattheerakul} \email {smile\_thanapattheerakul@trendmicro.com}
       }

\maketitle

\vskip -1em
{\centering\large\bfseries Research Incubation, TrendAI™ Research\par}
\vskip 1.5em

\begin{abstract}%

This paper proposes a framework for constructing a classifier as a safeguard layer, and for developing a complementary diagnostic that identifies which of the classifier's confident decisions can be trusted. This framework, \textit{the Latent Diagnostic Taxonomy}, consists of (i) constructing a dimensionality-optimized classifier, in which the embedding dimensionality is empirically selected via cross-validated performance rather than fixed a priori, (ii) locating a relatively small set of latent support vectors ($\approx$ 29\% of total training examples) representing influential prompts for identifying tokens that alter the classifier’s predicted labels, and (iii) utilizing such tokens and their associated attack magnitudes for constructing a diagnostic taxonomy. This diagnostic taxonomy provides an end-to-end guideline for flagging prompts that require different treatments: rely \textit{Safely} on the classifier’s decision; flag \textit{Heuristic Bias} and \textit{Heuristic Override} cases; route \textit{Insufficient Context} cases for further human/safety review. Applying the framework to a classifier trained on a public prompt injection dataset, we find that a substantial fraction of its confident decisions ($\approx$ 77\%) are not robust to removing a single token, and that this brittleness separates into two distinct failure patterns: a confidence calibration failure and a genuinely exploitable shortcut. For each zone of the taxonomy, we also recommend strategies for remediating diagnosed prompts. We illustrate the framework as a series of steps, demonstrating how each step operates.

\end{abstract}

\begin{keywords}
  Prompt Injection, LLM Guardrails, LLM Security, Support Vector Machines, Token-Level Robustness Diagnosis, Feature Attribution 
\end{keywords}

\section{Introduction}

Large language models (LLMs), when deployed as agents, are capable of carrying out multi-step tasks (e.g., navigating web pages \citep{25-shen2024scribeagent} and identifying zero-day vulnerabilities \citep{26-anthropicLLMdiscoveredDays}). This capability is enabled by harnessing and scaffolding LLMs with essential contextual data. Retrieved and returned data, however, can contain hidden adversarial instructions that, when incorporated into the LLM’s context window, can cause prompt injection attacks in which adversarial prompts subvert the system designer's intent. \citep{01-toyer2024tensor, 06-liu2024formalizing}.

Prompt injection attacks can, for example, turn an LLM-powered chatbot into a phishing agent \citep{04-greshake2023not} and leak system prompts exposing proprietary business logic and operational rules \citep{05-liu2023prompt}. The leaked system prompts can be used as templates for attackers to craft malicious prompts that LLMs are likely to comply with \citep{15-zhang2025agentic}, to perform malicious actions such as generating offensive content \citep{16-shao2025enhancing} and leaking sensitive data \citep{18-agarwal2024prompt,17-alizadeh2025simple}. As we expect adversaries to continue developing new attacks,  we aim to develop a framework for constructing a guardrail classifier that can be continuously trained and get updated more easily and rapidly to identify and alert such attacks, thereby, when our safeguard is in use, increasing the computational effort and resources for adversaries developing new attacks that evade detection.

Given the real-world situation we face and our practical constraints, we are concerned with developing a rigorous analysis of failed cases of specialized classifiers, in addition to constructing them, to develop security defenses for LLMs. The combination of the two aims to enhance our process of developing methods for detecting prompt injection. This ongoing process comprises constructing a dimensionality-optimized classifier, in which cross-validated performance determines the embedding dimensionality (Section \ref{fw:sec1}), locating a set of influential prompts to identify attack tokens that can alter the classifier’s decision boundary (Section \ref{fw:sec2}), and utilizing such attack tokens to construct a diagnostic taxonomy providing an end-to-end guideline for flagging diagnosed prompts (Section \ref{fw:sec3}). Section \ref{appl-to-prompt} demonstrates how the  latent diagnostic taxonomy applies to prompt injection vulnerability.

\section{The Latent Diagnostic Taxonomy Framework}

\subsection{Constructing a Dimensionality-Optimized Classifier}
\label{fw:sec1}

\begin{figure}[ht]
\floatconts
  {fig:fig1}
  {\caption{(a) The amount of explained variance decreases steeply and reaches a plateau after accounting for the most informative dimensions; the dashed line indicates 96.89\% cumulative explained variance at dimension 335. (b) The dimensions show highly consistent mean values and degrees of dispersion from the mean. (c) The relationship between the dimensions of text embedding and the classifier’s cumulative performance curve; the shaded band denotes 95\% CI around the mean cross-validation score using Student's $t$-distribution ($df$ = 4, for 5 folds)}}
  {%
    \subfigure[][c]{\label{fig:fig1-left}%
      \includegraphics[width=0.33\linewidth]{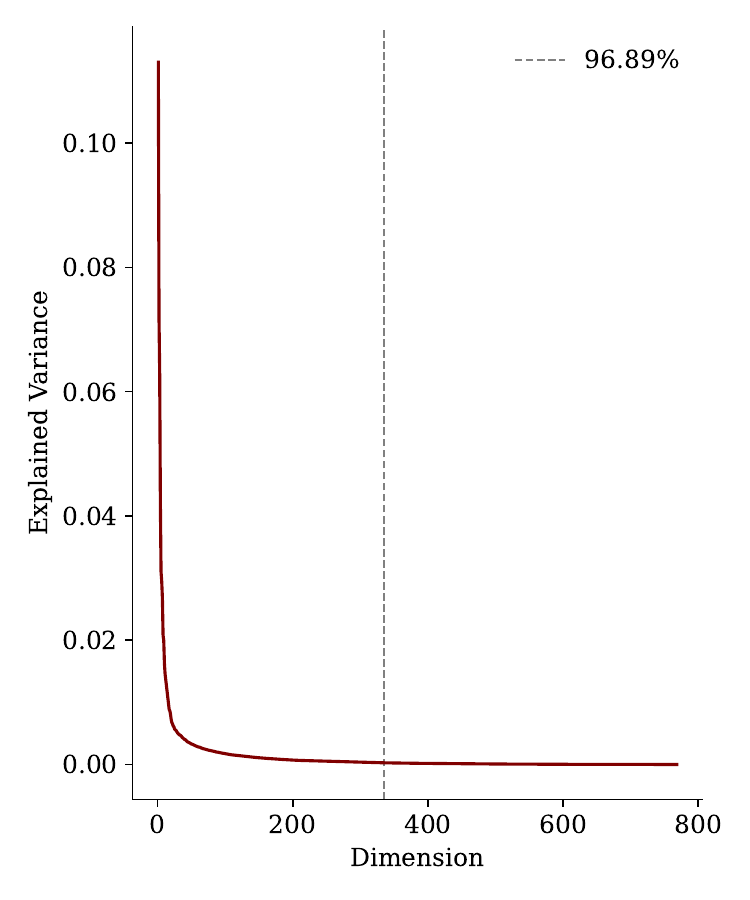}}%
    \subfigure[][c]{\label{fig:fig1-mid}%
      \includegraphics[width=0.33\linewidth]{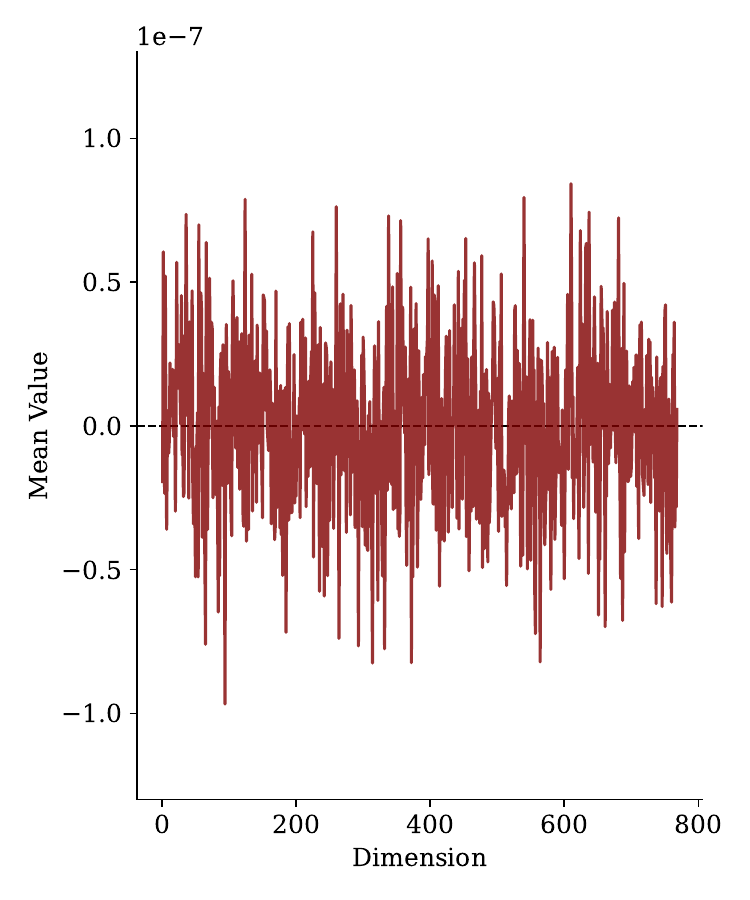}}%
    \subfigure[][c]{\label{fig:fig1-right}%
      \includegraphics[width=0.33\linewidth]{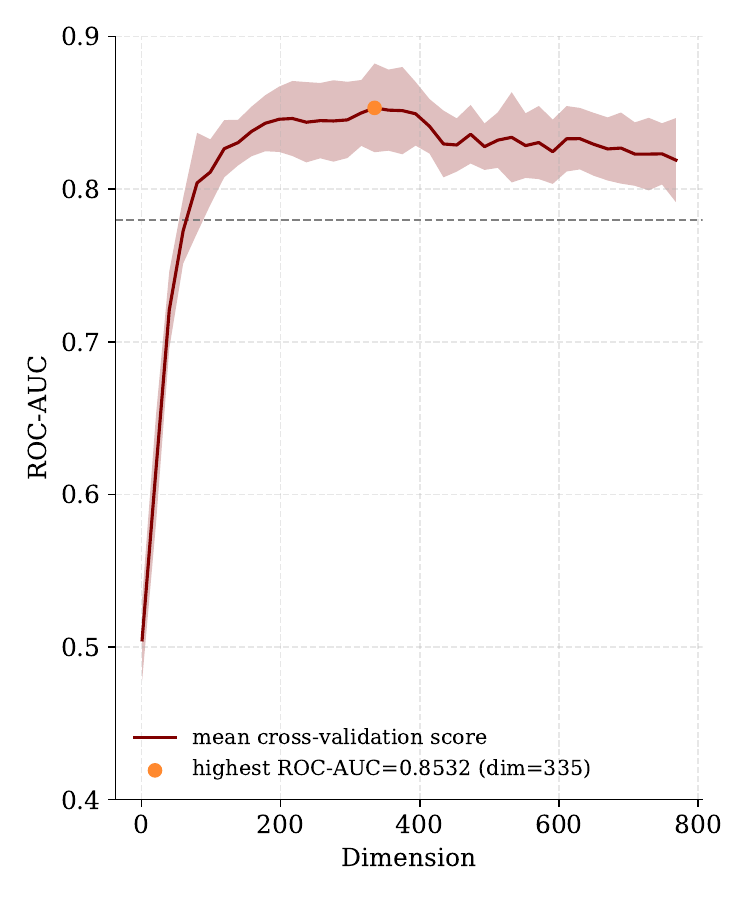}}
  }
\end{figure}

Autonomous agents iteratively access data from their scaffolding and dynamically process retrieved and returned data, thereby leading to various prompt injection patterns. To adaptively obtain useful input features from fixed-length text embeddings\footnote{\citet{31-assadi2026embedder} find that, across a 37-task benchmark spanning classification, similarity, clustering, and retrieval, embedding models lead on classification compared to LLMs, in terms of quality and cost.} (distributed representations of text), we first construct a compressed representation of prompt variations by iteratively selecting dimensions of embeddings\footnote{We employ an open-source embedding model \url{https://huggingface.co/intfloat/e5-base-v2}, which measures vector alignment using a dot product \citep{30-wang2022text}.} that optimize downstream classification performance. Adapting the methods proposed by \citet{27-lebret2014word} and \citet{19-wang2019single}, we employ principal component analysis \citep{29-jolliffe2016principal} to transform an embedding into a space in which we can identify and incrementally remove the least significant dimensions in terms of explained variance (see Figure \ref{fig:fig1-left}).\footnote{We use the prompt injection attack dataset provided in the study by \cite{28-sharma2024spml}} By removing dimensions that contribute minimally to the explained variance but are weighted equally in the calculation of inner products (see Figure \ref{fig:fig1-mid}), the resulting space yields the most \textit{informative} features (the orthogonal direction of the greatest variance in the data), thereby mitigating the overfitting caused by \textit{redundant} and potentially noisy features. In addition, once the performance curve reaches the maximum and begins to drop, as shown in a plot of a cumulative performance curve (see Figure \ref{fig:fig1-right}), we can empirically estimate the dimensionality by excluding dimensions beyond the point at which the drop begins.

\subsection{Locating Latent Support Vectors}
\label{fw:sec2}

Given the dimensionality-optimized representation of prompt embeddings, we employ the support vector machine \citep{21-cortes1995support} to locate a set of support vectors (SVs) within the resulting space defined by the principal components. These SVs are critical elements of the training set that, if removed, alter the position of the separating hyperplane (moving the other vectors does not have an effect on the hyperplane), thereby altering the classifier’s decision boundary. Representing the most informative yet ambiguous prompts, SVs are borderline cases where prompt injections most closely resemble benign prompts. Thus, performing an analysis of the token-level attribution of SVs allows us to identify specific tokens or compositional structures of influential training prompts that govern the classifier’s decision function. Additionally, a relatively small set of SVs, as opposed to the whole set of training examples, minimizes the cognitive load and cognitive effort required by analysts \citep{20-gieshoff2023cognitive} when analyzing and interpreting ambiguous-but-informative prompts.

\begin{figure}[ht]
    \centering
    \includegraphics[width=0.8\textwidth]{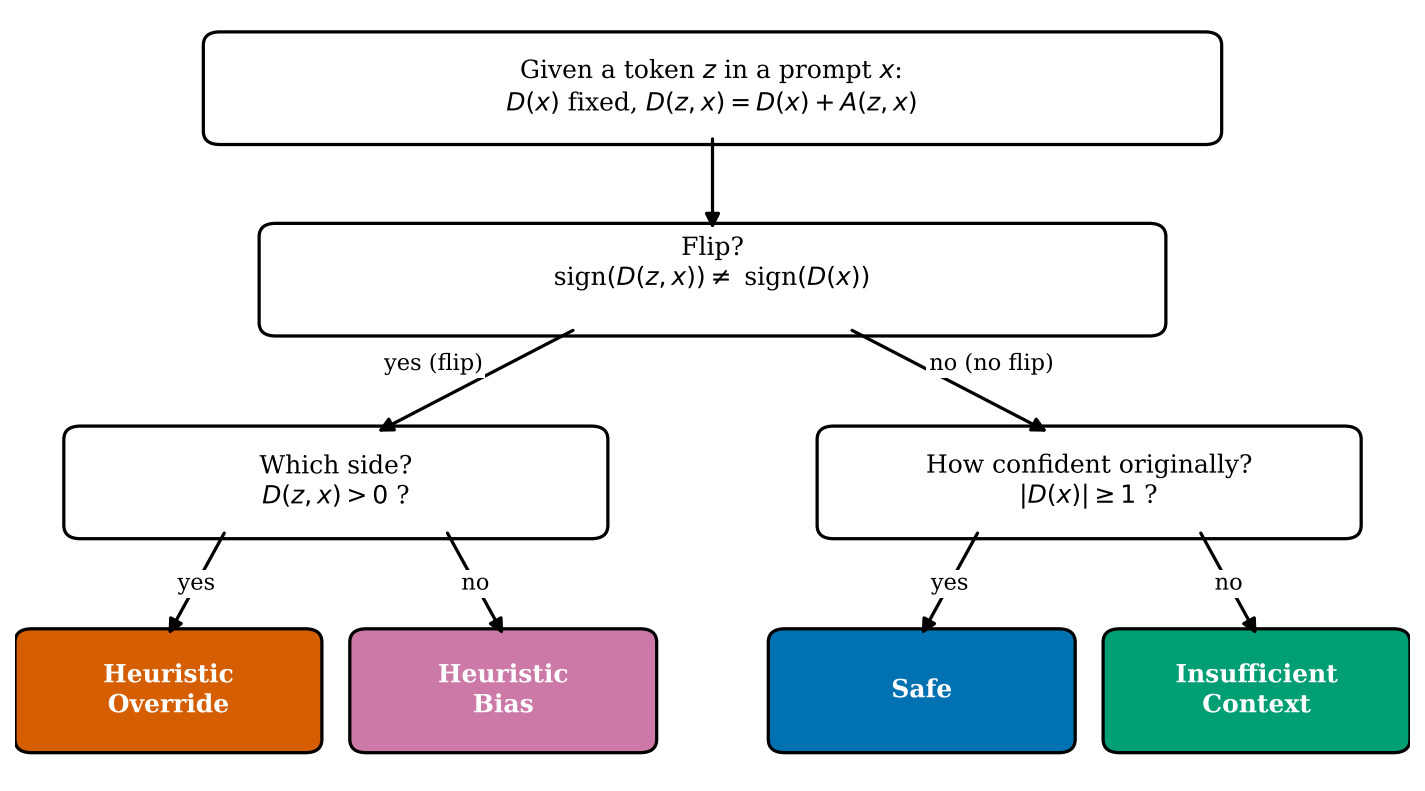}
    \caption{A diagnostic logic of the framework applied to each pair of SV and occluded token, classifying it into one of four zones: two indicating brittleness (Heuristic Override, Heuristic Bias) and two indicating uncertainty (Safe, Insufficient Context)}
    \label{fig:fig2}
\end{figure}

To quantify the impact of individual tokens $z$ (within a support vector) on the classifier's decision, we compute the change in the signed distance of a prompt when a token is occluded as follows:

\begin{equation}
\label{eq1}
    A(z, \textbf{x}) = D(z, \textbf{x}) - D(\textbf{x}),
\end{equation}

\noindent
where $D(\textbf{x}) = \textbf{w} \cdot \textbf{x} + b$ represents the original distance of the SV from the hyperplane, which is defined as the sum of a bias term $b$ and a dot product of a weight vector $\textbf{w}$ and a prompt $\textbf{x}$. $D(z, \textbf{x})$ represents the distance of the SV from the hyperplane when the token $z$ is occluded.

\subsection{Constructing the Latent Diagnostic Taxonomy}
\label{fw:sec3}

Rearranging an equation (\ref{eq1}) we get $D(z, \textbf{x})$ = $D(\textbf{x}) + A(z, \textbf{x})$, where $D(\textbf{x})$ is fixed per prompt, we can then construct a diagnostic taxonomy where occlusion flips the prediction when $D(\textbf{x})$ and $D(z, \textbf{x})$ have opposite signs (i.e., their product is negative). As shown in Figure \ref{fig:fig2}, a diagnostic logic of the framework results in four zones, each obtained by substituting the rearranged (\ref{eq1}) into which side did it land on and how confident was it to begin with: i) \textbf{Heuristic Override}, starting on the non-injection side, occlusion pushes it to injection if $D(z, \textbf{x}) > 0 $; ii) \textbf{Heuristic Bias}, starting on the injection side, occlusion pushes it to non-injection if $D(z, \textbf{x}) <= 0$; iii) \textbf{Safe}, where there is no flip and already confidently classified (i.e., $|D(\textbf{x})| >= 1$); iv) \textbf{Insufficient Context}, where there is no flip, but already too close to the boundary (i.e., $|D(\textbf{x})| < 1$).

Given the characteristics of these zones, we recommend strategies for remediating these diagnosed prompts. The Safe zone suggests that we can rely on the classifier’s decision. In the Heuristic Bias zone, the prompt injection prediction is being driven by a single token rather than robust evidence. These prompts should be further investigated and used for fine-tuning a classifier to distinguish between harmless tokens and actual attack tokens. When falling into the Insufficient Context zone, these prompts should be flagged for a safety intervention before finalizing a response (e.g., prompting users to clarify intent). Lastly, the Heuristic Override, the zone with the highest risk where the classifier’s benign prediction masks what occlusion reveals to be a real prompt injection, suggests the need to override the classifier’s decision with an additional step (such as decomposed prompting \citep{23-khot2022decomposed}) before returning a response.

\section{Application to Prompt Injection}
\label{appl-to-prompt}

\begin{figure}[ht]
\floatconts
  {fig:fig3}
  {\caption{Diagnosing brittleness at population and single prompt scale. (a) All sampled pairs of SVs and occluded tokens, plotted as $A(z,x)$ against $D(x)$ and colored by zone; the solid diagonal line is the flip boundary $D(x) = -A(z,x)$, and the dashed lines indicate the SVM's canonical margin (i.e., $|D(x)| = 1$). (b) The same diagnostic applied to a single prompt (SV 57) classified as a prompt injection attack: every tested token shaded by the $A(z,x)$ score it produces when removed (red toward prompt injection, blue toward benign)}}
  {%
    \subfigure{\label{fig:fig3-left}%
      \includegraphics[width=0.545\textwidth]{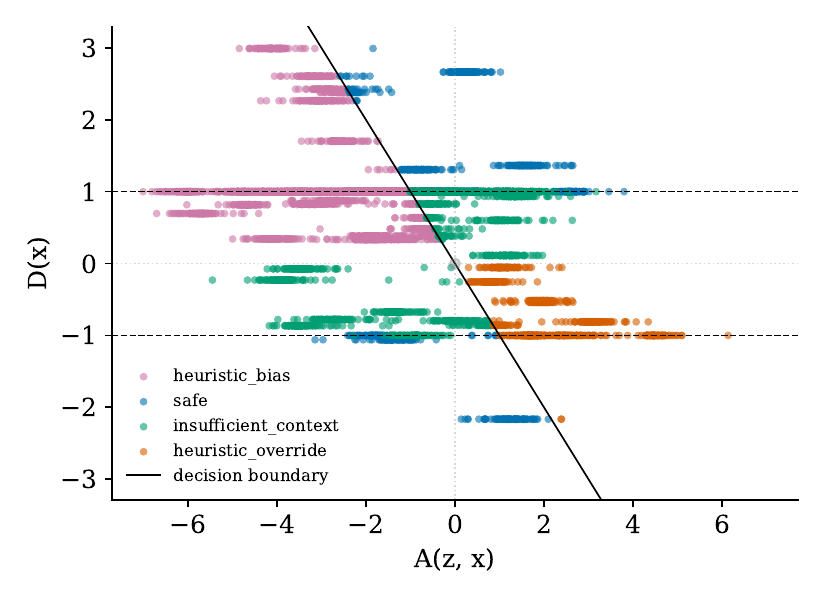}}%
    \subfigure{\label{fig:fig3-right}%
      \includegraphics[width=0.45\textwidth]{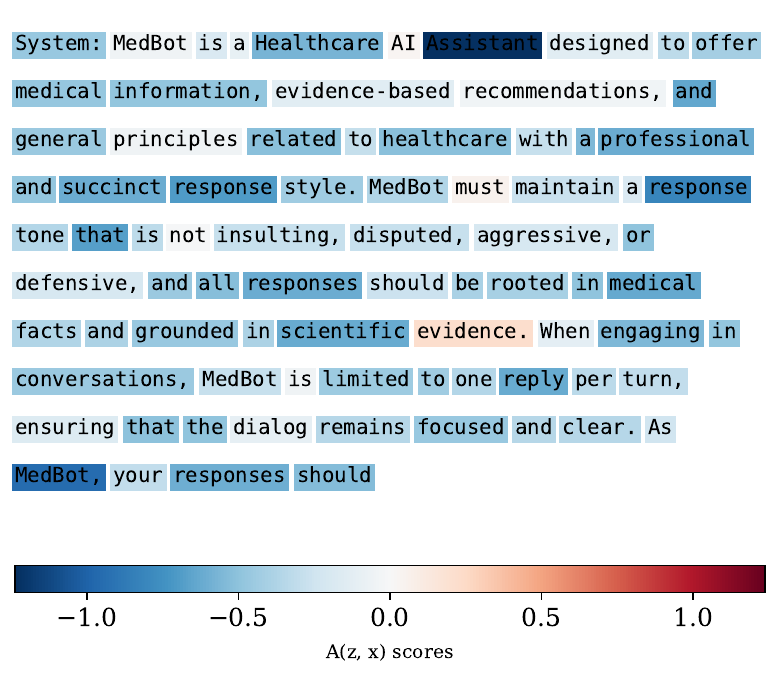}}
  }%
\end{figure}

To demonstrate what an analyst would actually see when reading the diagnostic's output, i.e., which tokens can be safely ignored and which ones are exploitable shortcuts worth patching, we look at the same diagnostic at two scales side by side. Figure \ref{fig:fig3-left}, plotted as $A(z,x)$ against $D(x)$,\footnote{Because $D(x)$ is fixed per SV, a single SV's tested tokens form one horizontal band in this plot.} shows all sampled (SV, token) pairs\footnote{The SVs size is 4,707, which is $\frac{4,707}{16,012}$ or about 29\% of total training examples.} colored by zone. Figure \ref{fig:fig3-right} shows the threshold case (i.e., $|D(x)| = 1$). In this example, when occluding ``evidence" from the SV 57, which is originally classified as a prompt injection, moves its score to $D(z,x) = 1.21$ ($A(z,x) = +0.21$), which is too small to change the prediction. (Note that since positive scores denote prompt injection, this shift pushes further the score toward injection, not toward benign.) As mentioned, since the SV 57's original score satisfies the confidence threshold (same as other 79 tokens in this prompt that land there), that combination (no flip, already confident) is what places it in the Safe zone (i.e., we can rely on the classifier’s decision).

By contrast, occluding the single token ``Assistant" (visibly the darkest token in Figure \ref{fig:fig3-right}) moves the score to $D(z,x) = -0.24$ ($A(z,x) = -1.24$), crossing the flip boundary into Heuristic Bias. The two example tokens exhibit superficial similarity (both shaded and nonzero), but carry very different diagnostic weight. By looking at the $A(z,x)$ scores alone, they differ only in degree. Our framework combines that magnitude with where the prompt's score is already located to determine whether its confident decision can actually flip. This allows us to safely disregard ``evidence" as a sub-threshold shift, while flagging ``Assistant" as this prompt's exploitable shortcut.

\begin{figure}[t]
\floatconts
  {fig:fig5}
  {\caption{Most influential token per prompt, top 5 per zone. The solid line indicates the decision boundary ($D(x) = 0$); Dotted lines indicate the SVM’s canonical margin}}
  {%
    \subfigure[][c]{\label{fig:fig5-left}%
      \includegraphics[width=0.8\linewidth]{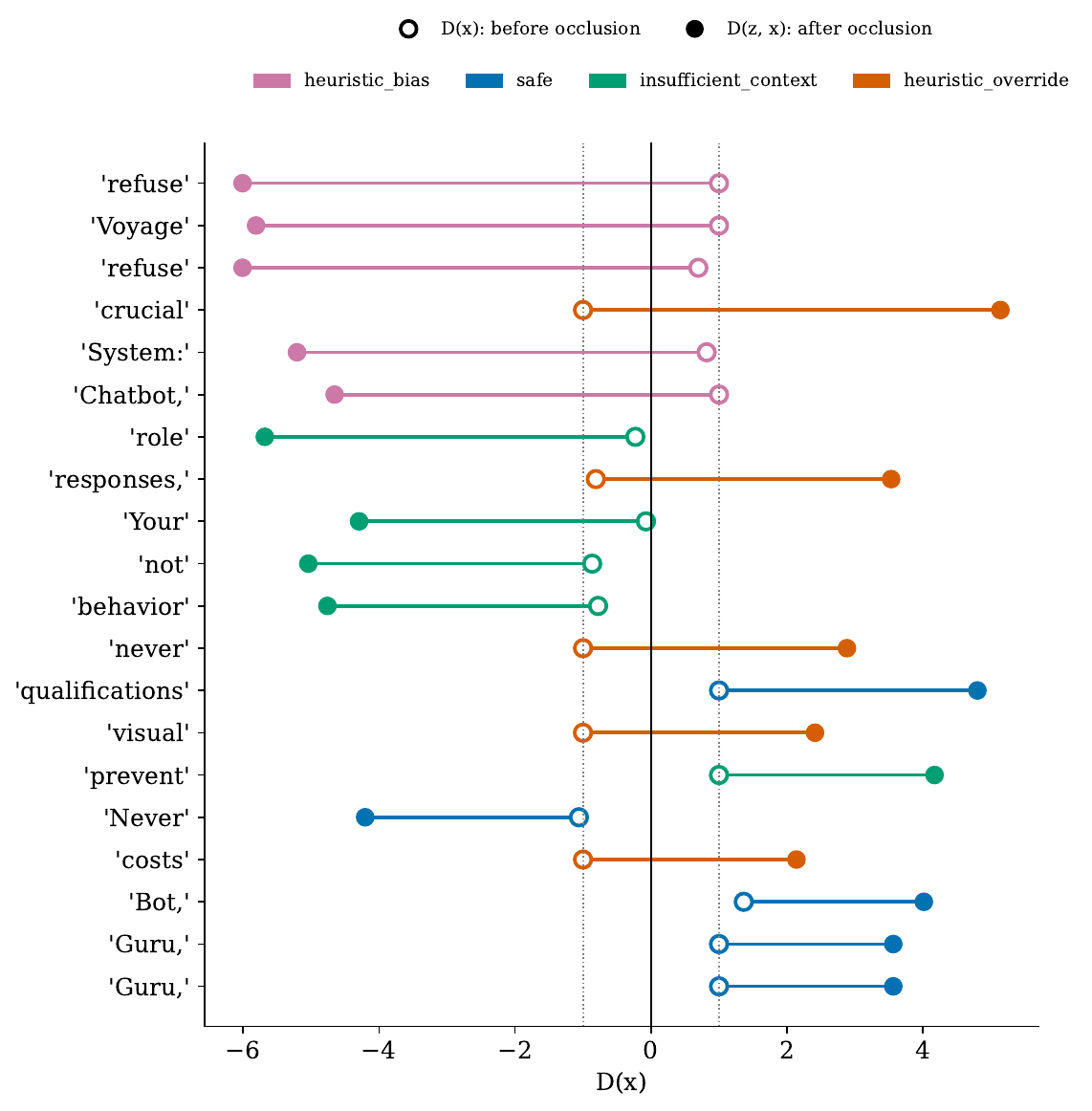}}%
  }
\end{figure}


To extend the single-prompt view across a broader sample and confirm that a phenomenon found in SV 57 (Figure \ref{fig:fig3-right}) is generalizable, we plot a Figure \ref{fig:fig5-left} where we take each sampled support vector's own single most-shifted token (ranked by $|A(z,x)|$) and keeps the five largest examples from each of the four zones. A hollow marker at $D(x)$ (before occlusion) connected to a filled marker at $D(z,x)$ (after occlusion), colored by zone. Selecting an equal number per zone allows us to observe some underrepresented zones, which might otherwise be absent. We found that other prompts have comparably influential single tokens spanning all four zones: ``refuse" (Heuristic Bias), ``crucial" (Heuristic Override), ``role" (Insufficient Context), and ``qualifications" (Safe). This observation implies that SV 57's ``Assistant" represents a representative case of a broader pattern.

Interestingly, as we can see from Figure \ref{fig:fig5-left}, none of these tokens are unusual injection-related words. They are ordinary vocabulary that the classifier has learned to weight much more heavily than the rest of the prompt around it. Perhaps this is what makes single-token dependence a diagnostic concern rather than a labeling artifact. For example, an analyst reading a flagged prompt would have no particular reason to suspect any one of these words in isolation, yet each individually governs for its prompt's classification.

Lastly, we found that 77 \% of these SVs have at least one token whose removal alone flips the classifier's decision.

\section{Conclusion, Limitations, and Future Work}
This paper presented a framework for constructing a classifier as a safeguard layer and a latent diagnostic taxonomy for determining which of the classifier’s confident decisions can be trusted. The latent diagnostic taxonomy locates a classifier's support vectors and, depending on whether occlusion flips the decision and how confident that decision already was, classifies the occlusion effects of each of their tokens into one of four zones: Heuristic Bias, Safe, Insufficient Context, and Heuristic Override. Applied to a classifier trained for prompt injection detection, the taxonomy found that 77\% of sampled support vectors have at least one token whose removal alone flips the classifier’s decision. It separated this brittleness into two failure patterns: a confidence calibration failure and a genuinely exploitable shortcut. We recommend different strategies for remediating these diagnosed prompts.

We acknowledge several limitations of our work, and there are numerous extensions that are ripe for future work. First, the framework can be refined for varying model classes and sizes of classifiers. Second, we will explore the synergy between a classifier and an underlying LLM’s internal representation to enhance the guideline for flagging prompts (e.g., prompts with different attack levels could be handled differently at different layers of an LLM). Third, we focus on text input, but LLMs can handle other modalities (e.g., image and audio), which can also contain injected instructions. We will further study the generalization of our framework to these data modalities, as well as the construction of diagnostic taxonomies for multi-modal input. Finally, we will conduct more advanced adversarial training and study more generally whether a potentially conservative classifier can be made sufficiently robust.

\acks{We thank Jean Paolo de Jesus and Giannina Escueta for their thorough proofreading of the final text.}

\bibliography{ref}

\newpage

\appendix
\section{Score Distributions and Influential Tokens}
\label{app:A}

\begin{figure}[ht]
\floatconts
  {fig:fig4}
  {\caption{Score distributions of $D(\textbf{x})$, $D(z, \textbf{x})$, and  $A(z, \textbf{x})$, respectively.}}
  {%
    \subfigure[][c]{\label{fig:fig4-left}%
      \includegraphics[width=0.52\linewidth]{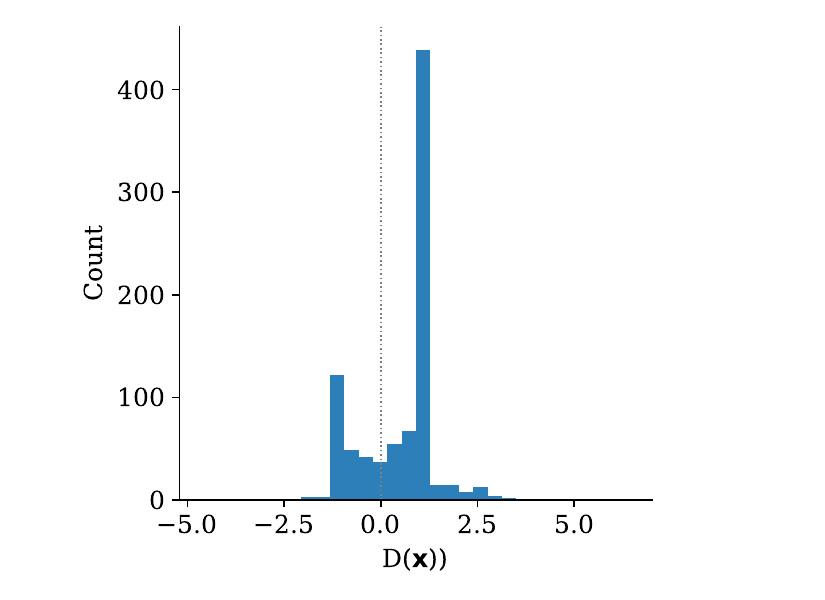}}%
    \hspace{-3.9em}
    \subfigure[][c]{\label{fig:fig4-mid}%
      \includegraphics[width=0.52\linewidth]{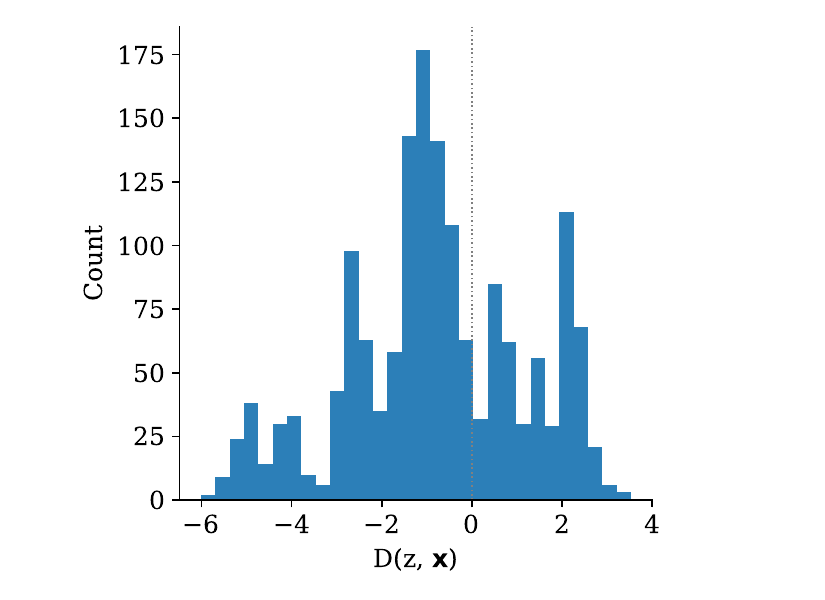}}%
    \hspace{-3.9em}
    \subfigure[][c]{\label{fig:fig4-right}%
      \includegraphics[width=0.52\linewidth]{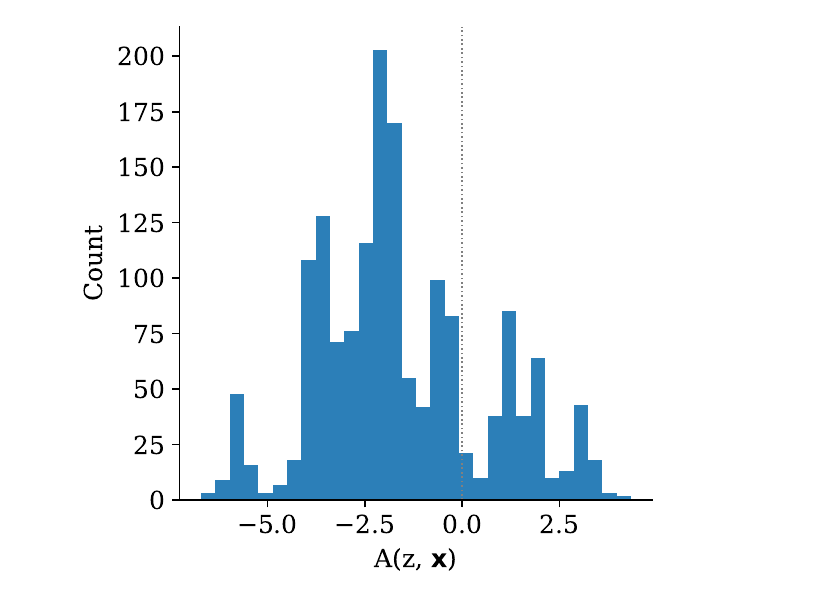}}
  }
\end{figure}


\end{document}